\documentclass{article}

    \usepackage[preprint]{neurips_2025}

\usepackage[utf8]{inputenc} % allow utf-8 input
\usepackage[T1]{fontenc}    % use 8-bit T1 fonts
\usepackage{hyperref}       % hyperlinks
\usepackage{url}            % simple URL typesetting
\usepackage{booktabs}       % professional-quality tables
\usepackage{amsfonts}       % blackboard math symbols
\usepackage{nicefrac}       % compact symbols for 1/2, etc.
\usepackage{microtype}      % microtypography

\usepackage{amsmath,bm}
\usepackage[table]{xcolor}
\usepackage{wrapfig}
\usepackage{algorithm}
\usepackage{multirow}
\usepackage{algpseudocode}
\usepackage{graphicx} 
\usepackage{subcaption}

\definecolor{citecolor}{HTML}{0071BC}
\hypersetup{colorlinks,linkcolor={red},citecolor={citecolor}}

\usepackage{xspace} % for proper spacing after commands
\newcommand{\eg}{\textit{e.g.}\xspace}
\newcommand{\ie}{\textit{i.e.}\xspace}
\newcommand{\etal}{\textit{et al.}\xspace}

\usepackage{comment}
\usepackage{newtxtext}
\usepackage{tikz}
\usepackage{pifont}
\newcommand{\red}[1]{{\color{red}#1}}
\newcommand{\cyan}[1]{{\color{cyan}#1}}
\newcommand{\yellow}[1]{{\color{orange}#1}}

\newcommand{\guid}{\text{guid}}
\newcommand{\inp}{\text{inp}}

\newcommand{\vae}{\text{vae}}
\newcommand{\gs}{\text{gs}}

\newcommand{\pse}{\text{pse}}
\newcommand{\den}{\text{den}}

\definecolor{top1}{RGB}{255,179,179}
\definecolor{top2}{RGB}{255,217,179}
\definecolor{top3}{RGB}{255,255,179}
\title{
Novel View Synthesis from A Few Glimpses via Test-Time Natural Video Completion
}
\author{%
\setlength{\tabcolsep}{30pt}
\begin{tabular}{@{}ccc@{}}
Yan Xu$^1$&
Yixing Wang$^1$&
Stella X. Yu$^{1,2}$\\
\end{tabular}\\
\setlength{\tabcolsep}{40pt}
\begin{tabular}{@{}lr@{}}
$^1$University of Michigan  & 
$^2$UC Berkeley \\
\end{tabular}\\
\texttt{\{yxumich,yixingw,stellayu\}@umich.edu}
}

\begin{document}

\maketitle
% \begin{wrapfigure}{r}{1\textwidth}
% Why feed-forward cannot replace 3D-GS
% - real-time rendering
% - view consistency

\begin{abstract}
Given just a few glimpses of a scene, can you imagine the movie playing out as the camera glides through it? That’s the lens we take on \emph{sparse-input novel view synthesis}, not only as filling spatial gaps between widely spaced views, but also as \emph{completing a natural video} unfolding through space.

We recast the task as \emph{test-time natural video completion}, using powerful priors from \emph{pretrained video diffusion models} to hallucinate plausible in-between views. Our \emph{zero-shot, generation-guided} framework produces pseudo views at novel camera poses, modulated by an \emph{uncertainty-aware mechanism} for spatial coherence.
These synthesized frames densify supervision for \emph{3D Gaussian Splatting} (3D-GS) for scene reconstruction, especially in under-observed regions. An iterative feedback loop lets 3D geometry and 2D view synthesis inform each other, improving both the scene reconstruction and the generated views.

The result is coherent, high-fidelity renderings from sparse inputs \emph{without any scene-specific training or fine-tuning}. On LLFF, DTU, DL3DV, and MipNeRF-360, our method significantly outperforms strong 3D-GS baselines under extreme sparsity.
Our project page is at \url{https://decayale.github.io/project/SV2CGS}. 
\end{abstract}

\section{Introduction}

% Humans can effortlessly imagine how a scene would look from unseen viewpoints by mentally filling in gaps---drawing on prior visual experience to interpolate what’s missing. We reinterpret novel view synthesis, a long-standing challenge in computer vision and graphics \cite{chen1993view, levoy1996light, paul1996modeling, thies2019deferred, mildenhall2021nerf, zhou2018stereo, single_view_mpi, kerbl20233d}, through a similar lens: as the task of completing a natural video from sparse camera views (Fig.~\ref{fig:teaser}).

\begin{figure}[!t]
    \centering
    \includegraphics[page=16, width=1\linewidth, trim={0cm 6cm 0cm 0cm},clip]{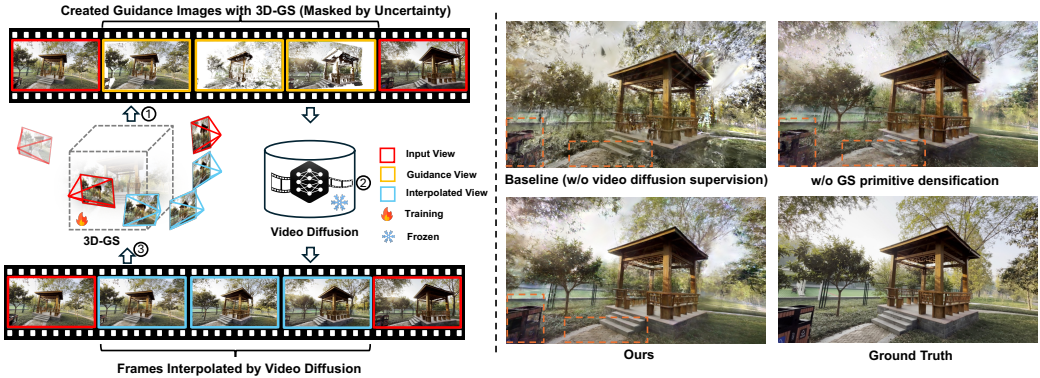}
    \vspace{-2em}
    \caption{
    We view sparse-input novel view synthesis as temporal-spatial completion of a natural-looking video.
    \textbf{Left:} Our generation-guided reconstruction pipeline. With the initialized 3D-GS from \red{sparse input views}, \ding{172} we create \yellow{guidance images} on interpolated poses and estimate their uncertainty, based on the currently optimized 3D-GS. \ding{173} Using both guidance images and their uncertainties, we modulate the diffusion score function to interpolate between sparse input views.  
    % Guided by the images, we interpolate the frames by modulating the score function of SVD in a zero-shot manner. 
    % The uncertainty estimation is incorporated during the reverse sampling for robustness. 
    \ding{174} The \cyan{interpolated views} are used to constrain 3D-GS optimization.  \textbf{Right:} With our generation-guide reconstruction, the under-observed regions in the inputs are enhanced by the views generated by the diffusion model. \vspace{-15pt}
    }
    \label{fig:teaser}
\end{figure}

Humans can effortlessly imagine how a scene appears from unseen viewpoints by mentally filling in gaps, by drawing on prior visual experience to infer what’s missing. Inspired by this ability, we reinterpret novel view synthesis -- a long-standing challenge in computer vision and graphics \cite{chen1993view, levoy1996light, paul1996modeling, thies2019deferred, mildenhall2021nerf, zhou2018stereo, single_view_mpi, kerbl20233d, lu2023urban} -- as the task of completing a natural video from sparse camera views (Fig.~\ref{fig:teaser}).
From this perspective, sparse-input novel view synthesis becomes analogous to recovering missing frames in a video captured along an unconstrained camera trajectory. This framing naturally invites the use of powerful generative priors learned from large-scale video data. 
In particular, pretrained video diffusion models~\cite{blattmann2023stable,yang2024cogvideox}, which are trained to synthesize coherent and realistic scene motions, offer a compelling tool for filling in plausible scene content between widely spaced views.

% Specifically, pretrained video diffusion models~\cite{blattmann2023stable,yang2024cogvideox}, trained to synthesize coherent and realistic scene movements, offer a compelling tool for filling in plausible scene content between widely spaced views.

Recently, NeRF~\cite{mildenhall2021nerf,barronmip,barron2023zip,muller2022instant} and 3D Gaussian Splatting (3D-GS)~\cite{kerbl20233d,yu2024mip,guedon2024sugar,huang20242d,lu2024scaffold} have significantly advanced novel view synthesis. Unlike NeRF, which represents scenes using an implicit function, 3D-GS models scenes explicitly with a set of 3D Gaussian primitives and renders images through efficient rasterization. 3D-GS achieves photorealistic rendering with substantially faster inference speed, making it a focal point of recent research interest.

However, synthesis from sparse inputs remains difficult. NeRF or 3D-GS methods typically rely on dense input views to accurately constrain the optimization process. In sparse-view settings, occlusions and geometric ambiguities~\cite{zhang2020nerf++} often lead to rendering artifacts and degraded quality. Recent efforts \cite{li2024dngaussian,zhu2025fsgs,chung2024depth,jain2021putting,wu2024reconfusion,xiong2023sparsegs} focus more on constrained camera paths (e.g., object-centric or forward-facing views).  In contrast, real-world image capture from walking with a handheld smartphone often produces widely spaced, unconstrained views with large occlusions and out-of-view regions (Fig.~\ref{fig:teaser}).

Motivated by the natural video completion perspective and strong priors in pretrained video diffusion models, we propose a {\bf zero-shot, generation-guided reconstruction} pipeline integrating video diffusion with 3D-GS. Our approach defines target camera trajectories between sparse input views and uses video diffusion priors to synthesize plausible intermediate pseudo-views. These views provide supervision to better constrain 3D-GS training, especially in the under-observed regions in the inputs.  
% and improve reconstruction quality.

To recover missing views along a natural video trajectory, we must generate images at specified camera poses. However, existing video diffusion models~\cite{blattmann2023stable,blattmann2023align,kong2024hunyuanvideo,wan2025} are typically conditioned only on the initial frame and produce uncontrolled camera motions. While recent methods~\cite{wang2024motionctrl,yu2024viewcrafter} introduce trajectory conditioning during training, they still lack guarantees of pose alignment at inference and rely heavily on datasets with 
% ground-truth 
camera parameters, limiting generalization and scalability.

We propose a novel \textit{uncertainty-aware modulation mechanism} that couples video diffusion with 3D Gaussian Splatting (3D-GS), enabling accurate, controllable frame interpolation under sparse-view settings. In this setup, 3D-GS provides a consistent 3D representation to guide view synthesis, while synthesized frames serve as pseudo supervision to further refine the 3D-GS model.

Fig.~\ref{fig:teaser} illustrates our overall workflow. Our method begins by initializing 3D-GS from sparse views. After initialization,  we interpolate camera poses between sparse inputs and create corresponding guidance images on the interpolated poses by inversely warping pixels from the nearest input view. The warping process is based on the depth maps rendered by the currently optimized 3D-GS. 
These guidance images are essential to maintaining the content and structural consistency during view interpolation, but may contain missing parts and artifacts due to imperfect 3D-GS depths and occlusion. We thus further model the uncertainty of these guidance images by assessing cross-view consistency in terms of photometry and geometry, and thereby focus the diffusion process more on correcting high-uncertainty regions, while keeping the reliable parts. 
Using both the guidance images and their associated uncertainties, we adaptively modulate the diffusion process to interpolate between the sparse views. 
The interpolated pseudo views are then added to the training set of 3D-GS. 
Furthermore, to improve the scene completeness for 3D-GS, we propose a \textit{Gaussian primitive densification module} to densify the 3D-GS point cloud in under-observed regions using these pseudo views as bridges.   
The process above is repeated iteratively to refine the 3D-GS reconstruction.

To summarize, our contributions are threefold:  
{\bf 1)} We propose a \textit{zero-shot, generation-guided} 3D-GS pipeline that leverages pretrained video diffusion models to improve novel view synthesis under sparse inputs, particularly in under-observed regions.  
{\bf 2)} We introduce an \textit{uncertainty-aware modulation} mechanism to integrate 3D-GS with video diffusion for controllable pseudo-view generation, and a \textit{Gaussian primitive densification} module to enhance scene completeness.  
{\bf 3)} Our method achieves state-of-the-art performance, with over 2.5 dB PSNR gain on DL3DV and strong results on LLFF and DTU, demonstrating robust generalization. While we primarily use Stable Video Diffusion~\cite{blattmann2023stable}, our framework is agnostic to the diffusion backbone and compatible with alternatives~\cite{yang2024cogvideox,kong2024hunyuanvideo}.

% \vspace{-20pt}
% \clearpage
\section{Related Work}
% \textbf{Novel View Synthesis based on Sparse Views}
% \cite{jain2021putting,wu2024reconfusion}

\textbf{Sparse-input Novel View Synthesis.} Sparse-input novel view synthesis aims to reconstruct a representation for generating novel views of a scene using a few input images. Although existing training-based methods, \ie NeRF \cite{mildenhall2021nerf} and 3DGS~\cite{kerbl20233d}, work well with dense inputs, their performance drops significantly with sparse views due to overfitting \cite{seo2023flipnerf, wang2023sparsenerf, niemeyer2022regnerf, deng2022depth, somraj2023simplenerf}.
Several recent works explore robust novel view synthesis under sparse inputs. One group~\cite{chen2022geoaug, niemeyer2022regnerf, jain2021putting, truong2023sparf, somraj2023VipNeRF, kwak2023geconerf, zhu2025fsgs,yin2024fewviewgs} focuses on imposing additional regularization on views deviating from the training views. For example, GeoAug~\cite{chen2022geoaug} randomly samples novel views around input frames and constrains rendering to match the input after view warping. Niemeyer \etal~\cite{niemeyer2022regnerf} introduce smooth depth regularization on unseen views.
SPARF~\cite{truong2023sparf}, GeCoNeRF~\cite{kwak2023geconerf}, and FewViewGS~\cite{yin2024fewviewgs} integrate multi-view correspondence and geometry loss into optimization. However, these methods do not address the fundamental issue of information deficiency in unobserved regions.

Another line of methods explores including priors from pre-trained neural networks~\cite{deng2022depth, wang2023sparsenerf, wynn2023diffusionerf,paliwal2025coherentgs,zhu2025fsgs,li2024dngaussian} for regularization. For example, Jain~\etal \cite{jain2021putting} leverage CLIP \cite{radford2021learning} features to provide semantic guidance. DSNeRF~\cite{deng2022depth} and SparseNeRF~\cite{wang2023sparsenerf} use depth regularization from pre-trained depth estimators on known views to guide optimization. 
% DiffusioNeRF~\cite{wynn2023diffusionerf} uses weight regularization and priors from diffusion models. 
More recently, FSGS~\cite{zhu2025fsgs} and DNGaussian~\cite{li2024dngaussian} extend the similar sprit to 3D-GS training.  However, these priors do not directly provide visual supervision for sparse-view NVS like the visual diffusion prior. 

\noindent\textbf{Novel View Synthesis with Diffusion Priors}.
To leverage visual priors for novel view synthesis, several lines of work have emerged. Liu \etal~\cite{liu2024deceptive} use diffusion models to generate pseudo-observations at unseen views, while Wu \etal~\cite{wu2024reconfusion} guide the diffusion process using a NeRF representation~\cite{yu2021pixelnerf} to synthesize novel views.

% To leverage the visual priors for novel view synthesis, Liu \etal \cite{liu2024deceptive} leverage diffusion models to generate pseudo-observations at unseen views.  Wu \etal \cite{wu2024reconfusion} generate novel views by guiding the trajectory of the diffusion process using a NeRF representation~\cite{yu2021pixelnerf}. 
To reduce the computational burden of fine-tuning diffusion models, Xiong \etal~\cite{xiong2023sparsegs} and Wang \etal~\cite{wang2025use} adopt Score Distillation Sampling (SDS)~\cite{poole2022dreamfusion} to extract external visual priors. However, these approaches rely on image-based diffusion models and thus fail to fully capture spatiotemporal correlations across views.
More recently, Liu \etal~\cite{liu3dgs} fine-tuned Stable Video Diffusion~\cite{blattmann2023stable} to provide view interpolation capability for guiding 3D-GS reconstruction. While this significantly improves performance, it requires substantial computational resources, limiting practical efficiency.

Despite progress in view-conditioned generative models~\cite{liu2023zero, wang2024dust3r, smart2024splatt3r, zhang2024monst3r}, existing methods are either object-centric~\cite{liu2023zero} or struggle to generate photorealistic views~\cite{smart2024splatt3r, zhang2024monst3r, wang2024dust3r}. Recent approaches~\cite{hou2024training, you2024nvs, wang2024motionctrl, yu2024viewcrafter} enable coarse camera motion control for video generation from a single frame but lack a consistent 3D representation, which compromises cross-view consistency and reproducibility.

% Recently, Liu~\etal~\cite{liu3dgs} proposed fine-tuning the Stable Video Diffusion~\cite{blattmann2023stable}, to endow the model with view interpolation ability, with which to guide 3D-GS reconstruction. While this improves performance significantly, it demands intensive computational resources, reducing practical efficiency.
% Despite progress in view-conditioned generative models~\cite{liu2023zero, wang2024dust3r, smart2024splatt3r, zhang2024monst3r}, these approaches remain either object-centric~\cite{liu2023zero} or struggle to generate photorealistic views~\cite{smart2024splatt3r, zhang2024monst3r, wang2024dust3r}.
% Recent works~\cite{hou2024training, you2024nvs, wang2024motionctrl,yu2024viewcrafter} enable coarse camera motion control for video generation from a single frame but lack a consistent 3D representation, compromising cross-view consistency and reproducibility. 
Consequently, how to effectively leverage zero-shot video diffusion priors for novel view synthesis is an important open challenge.
The concurrent work~\cite{zhong2025taming} is closely related to ours, but it depends on a video diffusion model trained with camera poses~\cite{yu2024viewcrafter}, and the code was not publicly available at the time of our submission. In contrast, our method can, in principle, be applied to any video diffusion model trained on raw videos, making it more broadly generalizable.

% \red{exp}

% Considering the recent view-conditioned generative model~\cite{liu2023zero, wang2024dust3r,smart2024splatt3r,zhang2024monst3r} for novel view generation is still either limited to object-centric~\cite{liu2023zero} or still have difficulties in generating photorealistic views~\cite{smart2024splatt3r,zhang2024monst3r,wang2024dust3r}, how to leverage the video diffusion priors in a zero-shot manner (without further training) for novel view synthesis has become a crucial problem. 

\begin{figure*}
    \centering
    \includegraphics[page=14, width=0.9\linewidth, trim={0cm 2cm 4cm 0cm},clip]{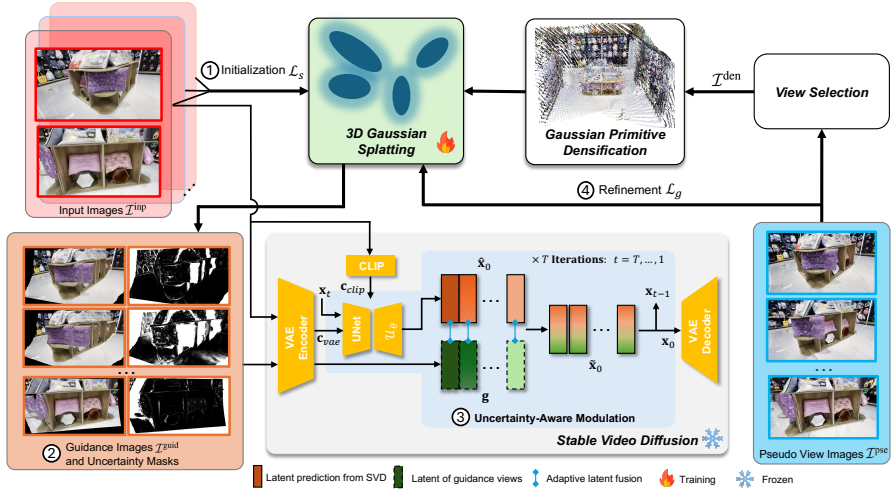}
    \caption{
    % To tackle the challenges of sparse-input novel view synthesis, we propose constraining the 3D-GS training process using pseudo views interpolated by a stable video diffusion (SVD) model. 
    \textbf{Overall framework}. After initializing 3D-GS from sparse input images (\ding{172}), \ding{173} we create guidance images (Sec.~\ref{sec:guidance_image}) and assess their uncertainties (Sec.~\ref{sec:uncertainty}) based on the current 3D-GS renderings.  
   \ding{174} The guidance images guide the diffusion process through the uncertainty-aware modulation (Sec.~\ref{sec:reverse_sample_modulation}). The diffusion process enhances high-uncertain regions while preserving reliable parts. \ding{175} The generated pseudo-view images are then used to densify the Gaussian primitives (Sec.~\ref{sec:gs_densification}) and to constrain the 3D-GS training (Sec.~\ref{sec:refinement}). 
   For illustration, we show pseudo-view generation from one image pair, though all pairs are processed sequentially in practice.
   % For clarity, we use one input image pair to illustrate the pseudo-view generation, whereas all the input images pairs would be processed one by one in our implementation.  
    % inversely warp the nearby input image to synthesize guidance images based on the rendered depth maps from 3D-GS and assess the uncertainty of the guidance images via cross-view consistency (Eq.~\eqref{eq:uncertainty}). 
    % To interpolate the frames without further training the SVD, we inversely warp the nearby input image to synthesize guidance images based on the rendered depth maps from 3D-GS. The uncertainty of guidance images is assessed by cross-view consistency (Eq.~\eqref{eq:uncertainty}). 
    % interpolate camera poses between adjacent input images and synthesize guidance images  based on the rendered depth by the current 3D-GS. whose uncertainties are estimated based on the current 3D-GS (Eq.~\eqref{eq:uncertainty}). 
    }
    \label{fig:framework}
    \vspace{-1em}
    
\end{figure*}

% \begin{comment}
\vspace{-0.5em}
\section{Preliminaries -- More Details in Appendix}\label{sec:prelim}
%\subsection{Preliminaries}\label{sec:prelim}
%\vspace{-0.5em}
% \subsection{3D Gaussian Splatting}
\noindent\textbf{3D Gaussian Splatting}
(3D-GS)~\cite{kerbl20233d} represents 3D scenes explicitly using Gaussian primitives, each defined by mean $\boldsymbol{\mu} \in \mathbb{R}^3$ and covariance $\boldsymbol{\Sigma} \in \mathbb{R}^{3\times 3}$:
% 3D Gaussian Splatting (3D-GS)~\cite{kerbl20233d} introduces an explicit representation of 3D scenes using a collection of 3D Gaussian primitives. 
% In this framework, each Gaussian is defined by a mean position $\boldsymbol{\mu} \in \mathbb{R}^3$ and a covariance matrix $\boldsymbol{\Sigma} \in \mathbb{R}^{3\times 3}$, which together determine its spatial distribution and shape in 3D space:  
$ G(\boldsymbol{x}) = \exp\left(-\frac{1}{2}(\boldsymbol{x} - \boldsymbol{\mu})^\top \boldsymbol{\Sigma}^{-1} (\boldsymbol{x} - \boldsymbol{\mu})\right) $. 
Each Gaussian also includes spherical harmonics coefficients $\boldsymbol{c}$ for view-dependent color and an opacity $\alpha$, enabling expressive appearance modeling.
% Each Gaussian carries additional attributes that define its appearance and transparency. These include spherical harmonics coefficients $\boldsymbol{c}$ for view-dependent color representation and an opacity parameter $\alpha$. This combination of spatial and appearance parameters allows for the flexible modeling of complex 3D scenes.
%
Rendering is performed efficiently via rasterization. After projecting Gaussians to the image plane, pixel colors are computed using alpha compositing:
% Given this explicit scene representation, 3D-GS's rendering process is quite efficient via rasterization.  After projecting the 3D Gaussians to the 2D image plane, the pixel colors can be obtained by blending their respective Gaussian colors and opacities through alpha compositing:
% \begin{equation}\label{eq:gs_render}
$C_{\text{pix}} = \sum_i \bm{c}_i \alpha_i \prod_{j=1}^{i-1}(1 - \alpha_j)$, 
% \end{equation}
% where $\bm{c}_i$ and $\alpha_i$ represent the color and opacity of each Gaussian, respectively. 
where $\bm{c}_i$ and $\alpha_i$ denote the color and opacity of the $i$-th Gaussian, respectively. For depth rendering, $\bm{c}_i$ is replaced by the z-buffer value. 
% To get the depth rendering, we can substitute $\mathbf{c}_i$ with the z-buffer of the corresponding Gaussian. More details are in the supplementary materials. 

% To render a scene, 3D Gaussians are projected onto the 2D image plane through a differentiable splatting process. Given a viewing transformation $\boldsymbol{W}$, each Gaussian is transformed into the camera coordinate system, and its covariance matrix is projected as:

% \begin{equation}
% \boldsymbol{\Sigma}^{\text{2D}} = \boldsymbol{J} \boldsymbol{W} \boldsymbol{\Sigma} \boldsymbol{W}^\top \boldsymbol{J}^\top,
% \end{equation}
% where $\boldsymbol{J}$ denotes the Jacobian of the projective transformation. The rendering process then computes the contribution of each Gaussian to each pixel, blending their respective colors and opacities through alpha compositing:
% \begin{equation}
% C_{\text{pix}} = \sum_i \bm{c}_i \alpha_i \prod_{j=1}^{i-1}(1 - \alpha_j).
% \end{equation}
% Here, $\bm{c}_i$ and $\alpha_i$ represent the color and opacity of each Gaussian, respectively. The final pixel color is obtained by blending overlapping Gaussian contributions, producing a smooth and high-quality rendering of the 3D scene. To get the depth rendering, we can substitute $\mathbf{c}_i$ with the z-buffer of the corresponding Gaussian. 

% \subsection{Diffusion Models}
{\bf Stable Video Diffusion} (SVD)~\cite{blattmann2023stable} is an image-to-video diffusion model that generates natural video conditioned on an input image. By default, generation starts from the given image and autonomously evolves, incorporating random camera movements and scene dynamics.
% In this section, we provide a brief overview of diffusion models~\cite{sohl2015deep}.
% % to facilitate the subsequent analyses. 
% For further details, we refer the reader to~\cite{song2019generative, song2020score}.

% \textbf{Forward Diffusion Process. }
Given a forward diffusion process expressed by $\mathrm{d}\mathbf{x} = f(t)\mathbf{x}\mathrm{d}t + g(t)\mathrm{d}\mathbf{w}$, where $\mathbf{x}$ is the noisy latent state at timestamp $t$, $\mathbf{w}$ denotes the standard Wiener process, and $f(t)$ and $g(t)$ are scalar functions, its reverse process ODE~\cite{song2020score} can be expressed as 
% \begin{equation}
% \label{eq:reverse1}
$\mathrm{d}\mathbf{x} = \left[f(t)\mathbf{x} - \frac{1}{2}g^2(\mathbf{x})\nabla_{\mathbf{x}} \log(q_t(\mathbf{x}))\right] \mathrm{d}t$.
% \end{equation}
% Since the noise of the diffusion is parameterized as i.i.d. Gaussian noises $\mathcal{N}(\bm{\epsilon}_t; \bm{0}, \sigma(t)\mathbf{I})$ with a variance of $\sigma(t)$, the above diffusion process can be calculated as
% \begin{equation}
% \label{eq:reverse3}
% d\bm{x} = \left[f(t)\bm{x} - \frac{1}{2}g^2(\bm{x})\frac{\bm{\mu}_{t} - \bm{x}}{\sigma^2(t)}\right] dt,
% \end{equation}
% where $\bm{\mu}_{t} = \mathcal{X}_{\bm{\theta}}(\bm{x}_t, t)$ is the estimated clean image from the noised image $\bm{x}_t$ at step $t$ by the denosing U-Net $\mathcal{X}_{\bm{\theta}}(\cdot,~\cdot)$\footnote{For simplicity, here we combine some parameterizations stemmed from EDM~\cite{karras2022elucidating} into~$\mathcal{X}_{\bm{\theta}}(\cdot)$.}.
In the case of the variance exploding (VE) diffusion~\cite{song2020score} adopted by Stable Video Diffusion (SVD)~\cite{blattmann2023stable}, 
%Eq.~\eqref{eq:reverse1} 
it can be simplified as:
% \begin{equation}
% \label{Eq:reverse}
$\mathrm{d}\mathbf{x} = \frac{\mathbf{x} - \hat{\mathbf{x}}_{0}}{\sigma_t} \mathrm{d}\sigma_t$, 
% \end{equation}
where the noise of the diffusion process is parameterized as Gaussian noise 
% $\mathcal{N}( \bm{0}, \sigma(t)\mathbf{I})$ 
with a variance of $\sigma_t$ and $\hat{\mathbf{x}}_0$ is the currently predicted clean video by the network based on the latent state at the previous step. 
% This formulation allows for an efficient and stable reverse diffusion process, enabling high-quality image and video generation.
In practice, we can obtain the estimated denoised sample 
$\mathbf{x}_{t-1}$ 
at the previous time step by discretizing the diffusion process above:
\begin{equation}
\label{eq:denoise_detail_main}
\mathbf{x}_{t-1} = \mathbf{x}_{t} + \frac{\mathbf{x}_t - \hat{\mathbf{x}}_0}{\sigma_t} (\sigma_{t-1}-\sigma_t).
\end{equation}

% \end{comment}

\vspace{-1em}
\section{Our Test-Time Optimization Approach to Novel View Synthesis}\label{sec:method}
\vspace{-0.5em}

% Our method integrates 3D scene reconstruction and 2D novel view synthesis through an iterative test-time optimization framework, leveraging 3D Gaussian Splatting and stable video diffusion for geometry and appearance, respectively.
We recast sparse-input novel view synthesis as a test-time natural video completion problem. To this end, we propose an iterative optimization framework that integrates 3D Gaussian Splatting with video diffusion priors to enforce geometric consistency and enhance visual fidelity. 

% Our method tackles sparse input novel view synthesis through an iterative test-time optimization framework, 
% leveraging 3D Gaussian Splatting and video diffusion priors for geometry and appearance, respectively.
% leverages 3D Gaussian Splatting 

% Given a set of sparse-view input images $\mathcal{I}^{\text{inp}}$, along with their poses, our goal is to reconstruct 3D Gaussian Splatting (3D-GS) representation~\cite{kerbl20233d} for novel view synthesis. To tackle this challenge, we introduce a zero-shot, generation-guided reconstruction pipeline that leverages a pretrained video diffusion model~\cite{blattmann2023stable}.The overall framework is shown in Fig.~\ref{fig:framework}.  

Given a few input views $\mathcal{I}^{\text{inp}}$ and their associated camera poses, we propose a zero-shot, generation-guided reconstruction pipeline that synthesizes novel views  
by leveraging a pretrained video diffusion model~\cite{blattmann2023stable} (Fig.~\ref{fig:framework}).
% The execution of our framework mainly consists of four steps: 
The framework consists of four main steps:
{\bf 1)} \textbf{3D-GS initialization} from the sparse input views; {\bf 2)}
% Guidance images creation 
\textbf{Guidance feature creation} 
(Sec.~\ref{sec:guidance_image}) and their \textbf{uncertainty estimation} via a cross-view consistency check(Sec.~\ref{sec:uncertainty}) based on the current 3D-GS; 
{\bf 3)} \textbf{Uncertainty-aware modulation} of the video diffusion model in generating pseudo views, conditioned on the guidance images and uncertainty masks (Sec.~\ref{sec:reverse_sample_modulation});
{\bf 4)} \textbf{Refinement of the 3D-GS} by densifying the Gaussian primitives using the generated pseudo-views (Sec.~\ref{sec:refinementa}).  
Steps \textbf{2)}–\textbf{4)} are iteratively performed to progressively improve both the 3D-GS representation and the quality of the diffusion model outputs.

% Steps \textbf{2}–\textbf{4} are iterated to progressively enhance both the 3D-GS and the diffusion model's outputs.

% Steps 2–4 can be iterated to enable mutual improvement between the 3D-GS and the diffusion model for better performance.

% The steps 2)-4)are iterated several times to make 3D-GS and diffusion model benefit each other. 

% Building on this framework, we iteratively optimize 3D-GS using these pseudo views and refine the pseudo view generation guided by the progressively improved 3D-GS model. Furthermore, by using these pseudo views as bridges, we densify the 3D-GS point cloud in under-observed regions to enhance rendering fidelity.
\vspace{-0.5em}
\subsection{Pseudo View Generation via Uncertainty-Aware Modulation}\label{sec:pseudo_view_generation}
\vspace{-0.5em}

Most off-the-shelf video diffusion models lack precise camera control due to the scarcity of datasets with known camera poses. To ensure broad applicability, we design our framework to be compatible with widely available models~\cite{blattmann2023stable,yang2024cogvideox} that are conditioned solely on a single image. Moreover, our approach is theoretically agnostic to variance-exploding diffusion backbones~\cite{song2020score}.

% Most off-the-shelf video diffusion models cannot be directly conditioned on camera parameters due to the scarcity of video datasets with ground-truth camera poses. To ensure broad applicability and generalization, we design our method to work with widely available video diffusion models~\cite{blattmann2023stable} that are conditioned only on the initial frame. This design choice allows our approach to be compatible with a wide range of existing video diffusion models~\cite{wan2025,kong2024hunyuanvideo} without requiring any model retraining or modification.

% As shown in Fig.~\ref{fig:framework}, 

The modern video diffusion model~\cite{blattmann2023stable} usually extracts CLIP~\cite{radford2021learning} features $\mathbf{c}_{\text{clip}}$ from the input frame $I{^\text{inp}}$ to inform the U-Net of the scene's overall appearance and layout. Simultaneously, the frame is encoded by a VAE encoder to produce contextual features $\mathbf{c}_{\vae}$, which are injected via classifier-free guidance to maintain consistency with the reference frame.
 At each denoising timestep $t$, the model denoises a latent video representation $\mathbf{x}_t \in \mathbb{R}^{N \times C \times H \times W}$ using a U-Net $\mathcal{U}_{\bm{\theta}}(\mathbf{x}_t; \mathbf{c}_{\text{clip}}, \mathbf{c}_{\text{vae}}, t)$, where $N$, $C$, $H$, $W$ are the number of frames,  feature and spatial dimensions of the latent, respectively. The U-Net predicts a clean latent $\hat{\mathbf{x}}_0$ from $\mathbf{x}_t$ to update $\mathbf{x}_t$ with Eq.~\eqref{eq:denoise_detail_main}, which direct $\mathbf{x}_t$ toward $\hat{\mathbf{x}}_0$. 
 % to iteratively update
% During reverse sampling, 
% $\mathbf{x}_t$ is iteratively updated using Eq.~\eqref{eq:denoise_detail}, guiding it toward $\hat{\mathbf{x}}_0$. 
The final denoised latent,  $\mathbf{x}_0$, is decoded by the VAE decoder into a video clip. 

% As shown in Fig.~\ref{fig:framework}, the modern video diffusion model~\cite{blattmann2023stable} extracts CLIP features $\mathbf{c}_{\clip}$ from the input frame $I_{\inp}$ to inform the U-Net of the scene's overall appearance and layout. In addition, the input frame is encoded by a VAE encoder to produce contextual features $\mathbf{c}_{\vae}$, which are injected via classifier-free guidance to help maintain consistency with the reference frame.
% After image feature encoding, the diffusion model uses a U-Net $\mathcal{U}_{\bm{\theta}}(\mathbf{x}_t; \mathbf{c}_{\clip}, \mathbf{c}_{\vae}, t)$ to denoise a latent representation $\mathbf{x}_t \in \mathbb{R}^{N \times H \times W}$ at each timestep $t$, where $N$ is the number of video frames and $H \times W$ the spatial resolution. 
% At each timestep, the U-Net predicts a clean latent $\hat{\mathbf{x}}_0$ from the currently denoised $\mathbf{x}_t$, which can be decoded by the VAE decoder into a video clip. For reverse sampling, the latent $\mathbf{x}_t$ is iteratively updated using Eq.~\eqref{eq:denoise_detail}, effectively guiding it toward $\hat{\mathbf{x}}_0$.

Our method draws inspiration from diffusion-based image editing techniques~\cite{mengsdedit,yu2023freedom,bansal2023universal,xu2024inversion}, particularly SDEdit~\cite{mengsdedit} for its efficiency.
% Our method is inspired by image editing techniques using diffusion models~\cite{mengsdedit,yu2023freedom,bansal2023universal,xu2024inversion}, particularly SDEdit~\cite{mengsdedit} due to its efficiency.
Specifically, we propose to modify the original clean latent prediction $\hat{\mathbf{x}}_0$ 
% (predicted by the diffusion U-Net) 
using the guidance feature $\mathbf{g}\in \mathbb{R}^{N \times C \times H \times W}$ extracted from the guidance images by the VAE encoder. 
This modification is formulated as an optimization problem applied to each frame $i$:
% This modification is formulated as an optimization problem for each frame $i$:
\begin{equation}\label{eq:fusion_solution}
% \widetilde{\mathbf{x}}_{0}^{i} = \operatorname*{arg~min}_{\mathbf{x}} \| \mathbf{x} -  \hat{\mathbf{x}}_{0}^{i}\|_2^2 + \gamma_{t, i} \| \mathbf{x} - {\mathbf{g}}^{i} \|_2^2,
\widetilde{\mathbf{x}}_{0}[i] = \operatorname*{arg~min}_{\mathbf{x}} \| \mathbf{x} -  \hat{\mathbf{x}}_{0}[i]\|_2^2 + \mathbf{\gamma}_{t,i} \| \mathbf{x} - {\mathbf{g}}[i] \|_2^2, 
\end{equation}
where index $[i]$ denotes the $i$-th frame channel corresponding to the $i$-th frame of the generated video, and $\gamma_{t,i} > 0$ is a weighting term that controls the influence of the guidance feature. 
% where $\mathbf{g}[i]$ denotes the $i$-th frame channel of the encoded guidance feature map corresponding to the $i$-th frame of the generated video, and $\gamma_{t,i} > 0$ is a weighting term that controls the influence of the guidance feature. 
% where $\mathbf{g}_{i}$ is the VAE-encoded feature for guidance image associated with the $i$-th generation frame of the video output 
% $I^{g}_{i}$ 
% We use the subscript $t$ to denote the diffusion timestep and use bracketed $i$ to denote the channel index of the feature map (corresponding to the $i$-th generation frame).  

The remaining problems are \textbf{1)}
% how to create the guidance images to 
how to get the proper feature map $\mathbf{g}$ to guide the diffusion model in generating views of desired poses (Sec. \ref{sec:guidance_image})
and \textbf{2)} how to control $\gamma_{t,i}$ to achieve adaptive modulation (Sec. \ref{sec:uncertainty}-\ref{sec:reverse_sample_modulation}).

% \vspace{-0.5em}
% \subsubsection{Guidance Image Creation} \label{sec:guidance_image}
\subsubsection{Guidance Feature Creation} \label{sec:guidance_image}
% \vspace{-0.5em}
% Unlike the stroke guidance image adopted by the image editing scenarios, in our setting, the guidance images on specific poses need to geometrically align well with the underlying scene to help keep the fidelity of the view generation.   
% Our high-level idea is to utilize the priors from the video diffusion model to generate the occluded or missing regions from the sparse inputs. To achieve this, we need to create guidance features that are geometrically aligned with the view we want to generate. To this end, 
The core idea of our approach is to exploit video diffusion priors to infer occluded or missing content from sparse input views. This requires constructing guidance features that are geometrically aligned with the desired target view.
% To create guidance features that are geometrically aligned with target views, 
To this end, a simple strategy is to render the target views from the current 3D-GS and encode them with the diffusion model’s VAE encoder, thereby maintaining 3D consistency.
However, this often yields low-fidelity results, as 3D-GS may produce inaccurate color renderings at novel poses during training. 

To resolve this, instead of using the 3D-GS to render color images, we create guidance images by inversely warping pixels from their nearest input view, using depth maps rendered by 3D-GS.
Concretely, to construct the guidance image $I^\guid_i$ for the $i$-th video frame, we first project each pixel $\mathbf{p}\in I^\guid$ into the nearest input view $I^\inp\in\mathcal{I}^\inp$, using the rendered depth map $D_i^\guid$, camera intrinsics $\mathbf{K}$, and camera poses $\mathbf{P}^\inp \in \mathbb{SE}(3)$ (input view) and $\mathbf{P}_i^\guid \in \mathbb{SE}(3)$ (guidance view), to get its corresponding pixel $\mathbf{q}$ in the input image:
% is computed as:
\begin{equation}\label{eq:forward_proj}
\mathbf{q} = \mathbf{K} \mathbf{P}^\inp (\mathbf{P}_i^\guid)^{-1} D_i^\guid(\mathbf{p}) \mathbf{K}^{-1} \mathbf{p}.
\end{equation}
We fill pixel $\mathbf{p}$ with the color of pixel $\mathbf{q}$ to obtain the guidance image $I^\guid_i$. The set of guidance images is denoted as $\mathcal{I}^{\guid} = \{I^{\guid}_i\}_{i=1}^{N}$, where $N$ is the length of the video clip generated by the video diffusion model in a single pass. 
% generated frames in a single forward pass of the diffusion model. 
The VAE encoder will encode these guidance images to have the corresponding guidance feature maps $\mathbf{g}$ to guide the diffusion process via Eq.~\eqref{eq:fusion_solution}.

% The warping process can be expressed as: 
% \begin{equation}\label{eq:forward_proj}
% % \mathbf{q} = \mathbf{K}\mathbf{P}_{inp \leftarrow i } {D}_i(\mathbf{p})\mathbf{K}^{-1}\mathbf{p}, 
% % \mathbf{q} = \mathbf{K}\mathbf{P}_{inp \leftarrow i } {D}_i^{\guid}(\mathbf{p})\mathbf{K}^{-1}\mathbf{p}, 
% % \mathbf{q} = \mathbf{K}\mathbf{P}^{\inp} (\mathbf{P}_{i}^{\guid})^{-1} {D}_i^{\guid}(\mathbf{p})\mathbf{K}^{-1}\mathbf{p}, 
% \mathbf{q} = \mathbf{K}\mathbf{P}^{\inp} (\mathbf{P}_{i}^\guid)^{-1} {D}_i^\guid(\mathbf{p})\mathbf{K}^{-1}\mathbf{p}, 
% % \mathbf{q} = \mathbf{K}\mathbf{P}^{inp\leftarrow guid}_i  (\mathbf{p})\mathbf{K}^{-1}\mathbf{p}, 
% \end{equation}
% where $\mathbf{p}$ is a pixel we want to create in the guidance image $I^\guid_i$ for the $i$-th video frame, and $\mathbf{q}$ is the corresponding pixel in the nearest input image {$I^\inp\in \mathcal{I}^{\inp}$}.
% Given the rendered depth map $D_i^{\guid}$ of the guidance view, 
% camera intrinsics $\mathbf{K}$, and the camera poses $\mathbf{P}^\inp$ for the input view and  $\mathbf{P}_{i}^{\guid}$ for the guidance view, 
% we compute $\mathbf{q}$ for each $\mathbf{p}$ and apply nearest-neighbor interpolation to construct the guidance image $I^\guid_i$. We denote these created guidance images as: $\mathcal{I}^{\guid}=\{{I^{\guid}_i}\}_{i=1}^{N}$, where $N$ is the total number of frames to interpolate. 

\begin{figure}[tb]
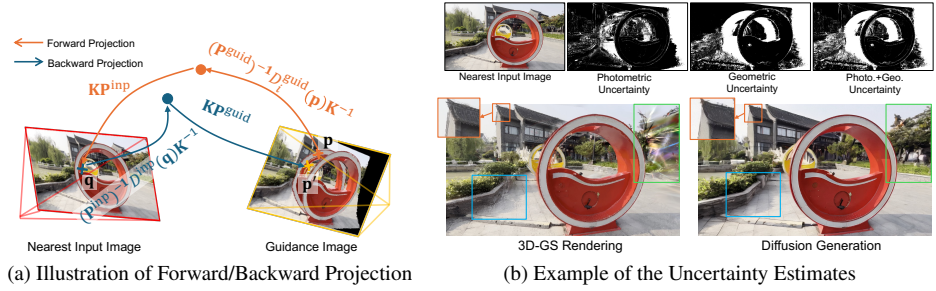

    \centering
     \resizebox{0.9\linewidth}{!}{
    \begin{subfigure}[b]{0.44\textwidth}
         \includegraphics[page=15, width=1\linewidth, trim={1cm 11cm 19cm 0cm},clip]{fig/diagrams.pdf}
    \vspace{-1.5em}
        \caption{Illustration of Forward/Backward Projection}
        \label{fig:cross_view_check}
    \end{subfigure}
    \hfill
    \begin{subfigure}[b]{0.55\textwidth}
         \includegraphics[page=3, width=1\linewidth, trim={0cm 6cm 9cm 0.5cm},clip]{fig/diagrams.pdf}
    \vspace{-1.5em}
        \caption{Example of the Uncertainty Estimates}\label{fig:geo_photo_unceratinty}
        \label{fig:figure1}
    \end{subfigure}
    }
    % \vspace{-0.5em}
    \caption{\textbf{Cross-view consistency is evaluated through the forward and backward projections} shown in (a) to estimate the uncertainty of the generated guidance image. As illustrated in (b), regions exhibiting poor cross-view consistency (regions in the boxes) are identified as high-uncertainty areas (brighter), which are subsequently refined by the video diffusion model. 
    % As shown in (b),  unreliable regions in the guidance image are effectively identified. These regions are subsequently refined by the video diffusion model. 
    % Illustration of the forward and backward projection process used in the cross-view consistency check.
% (b) Example of the uncertainty maps evaluated for a generated pseudo-view. By combining photometric and geometric uncertainty estimates, unreliable regions in the guidance image are effectively identified. These regions are subsequently refined by the video diffusion model.
    % \red{TODO: add guidance image}
    }
    \vspace{-1em}
\end{figure}

\vspace{-0.5em}
\subsubsection{Uncertainty Evaluation from Cross-View Consistency}\label{sec:uncertainty}
\vspace{-0.5em}
The constructed guidance images well preserve scene content and structure by adhering to strict multi-view geometric constraints imposed by the 3D-GS representation. However, because 3D-GS is imperfect during training, especially in under-observed regions, the guidance images may contain missing content or artifacts.
% These guidance images maintain the content and structural consistency of the scene, given the strict multi-view geometric constraint from the 3D-GS representation. However, the 3D-GS during training is imperfect, especially in the under-observed regions in the inputs, which may lead to missing parts and artifacts in the guidance images. 
To assess the reliability of guidance images, we introduce a strict cyclic consistency check, as illustrated in Fig.~\ref{fig:cross_view_check}.
Specifically, in the forward pass, we project each pixel $\mathbf{p}$ in the guidance image to its corresponding pixel $\mathbf{q}$ in the nearest input image 
% (the very image used to create this guidance image), 
using Eq.~\eqref{eq:forward_proj}.
We then perform a backward projection from $\mathbf{q}$ to the guidance view using the depth map $D^{\inp}$ rendered by 3D-GS from the nearest input view:
% To evaluate the uncertainty of guidance images, we introduce a strict cyclic consistency check mechanism, as illustrated in Fig.~\ref{fig:cross_view_check}. 
% % based on the fact that if a pixel is poorly constrained, its geometric/photometric rendering from 3D-GS is generally erroneous. 
% Specifically, in the forward pass, we follow the guidance image creation procedure to project the pixel $\mathbf{p}$ in the guidance image to the nearest input view with Eq.~\eqref{eq:forward_proj}, getting its corresponding pixel $\mathbf{q}$ defined by the current 3D-GS. 
% Then, we project $\mathbf{q}$ back to the guidance image space with the reference image's depth map $D_{ref}$ rendered by 3D-GS: 
% $\mathbf{p}' = \mathbf{K}\mathbf{P}_{inp\leftarrow i}^{-1} {D}^{\inp}(\mathbf{q})\mathbf{K}^{-1}\mathbf{q}$. 
$\mathbf{p}' = \mathbf{K} \mathbf{P}_{i}^{\guid}(\mathbf{P}^{\inp})^{-1} 
  {D}^{\inp}(\mathbf{q})\mathbf{K}^{-1}\mathbf{q}$. 
The uncertainty at pixel $\mathbf{p}$ is then quantified by evaluating both geometric and photometric consistency:
\begin{equation}\label{eq:uncertainty}
   % U_i(\mathbf{p}) = 1 - \exp\left( - \frac{ ||\mathbf{p}-\mathbf{p}'||_2^2 }{s_1} -\frac{||I^{\gs}_i(\mathbf{p})- I^{\inp}(\mathbf{q})||_2^2}{s_2} \right), 
   U_i(\mathbf{p}) = 1 - \exp\left( - \frac{ ||\mathbf{p}-\mathbf{p}'||_2^2 }{s_1} -\frac{||I^{\gs}_i(\mathbf{p})- I^{\inp}(\mathbf{q})||_2^2}{s_2} \right), 
\end{equation}
where $I^{\gs}_i$ is the 3D-GS rendered image from the view of the $i$-th guidance image, $I^{\inp}$ denotes the nearest input image, and $s_1$, $s_2$ are bandwidth parameters controlling the sensitivity to geometric and photometric discrepancies.
% where $I_i$ is the rendered color image by 3D-GS at the guidance image's pose and $I_{ref}$ denotes the reference image. $s_1$ and $s_2$ are two bandwidth parameters to balance the geometric and photometric terms. 
% If the 3D-GS is well constrained at pixel $\mathbf{p}$ and no occlusion exists, 
% the $I_{ref}(\mathbf{q})$ should have a color similar to the 3D-GS rendering $I_i(\mathbf{p})$ and the source pixel $\mathbf{p}$ and the back-projected $\mathbf{p}'$ should stay close to the source pixel $\mathbf{p}$, leading to lower uncertainty, otherwise, the uncertainty estimation would increase according to Eq.~\ref{eq:uncertainty}. 
If the 3D-GS is well constrained at pixel $\mathbf{p}$ and no occlusion is present, the image pixel color $I^{\inp}(\mathbf{q})$ should closely match the color of the 3D-GS rendering $I_i^{\gs}(\mathbf{p})$, and the back-projected position $\mathbf{p}'$ should lie near the original $\mathbf{p}$. This results in low uncertainty. Otherwise, discrepancies in color or geometry increase the uncertainty, as captured by Eq.~\eqref{eq:uncertainty}.

\vspace{-0.5em}
\subsubsection{Uncertainty-Aware Modulation}\label{sec:reverse_sample_modulation}
\vspace{-0.5em}

Using the uncertainty map, we define $\gamma_{t,i}$ for each pixel in Eq.~\eqref{eq:fusion_solution} as: 
\begin{equation}\label{eq:gamma}
    \gamma_{t,i}(\mathbf{p}) = \left\{
    \begin {aligned}
         % & 0 \quad & U_i(\mathbf{p}) > \delta~\mathrm{or}~t<\tau \\
         % & 1 \quad & otherwise % 
         & 0 \quad  & U_i(\mathbf{p}) > \delta~\mathrm{or}~t<\tau \\
         % &  (1-U_i(\mathbf{p})) \quad & U_i(\mathbf{p})<\delta~and~t\geq\tau \\
         % &  \frac{1}{U_i(\mathbf{p})} \quad & otherwise % 
         &  {1}/({U_i(\mathbf{p})} +\epsilon )\quad & \text{otherwise}% 
    \end{aligned}
\right.,
\end{equation}
where $\delta$ and $\tau$ are threshold hyperparameters and $\epsilon$ is a small constant to avoid division by zero. 
The threshold $\tau$ is determined by the overall uncertainty of frame $i$, measured by 
% $\tau$ is defined as a linear function defined on the sum of all the pixel uncertainty within frame $i$: 
$\tau= \frac{k}{HW}\sum_{\mathbf{p}}(U_i(\mathbf{p}))+b$, with $k$ and $b$ as tunable coefficients. % (see supplementary materials for details).
% Please refer to the supplementary materials for details. 
This ensures that in uncertain regions, the optimization in Eq.~\eqref{eq:fusion_solution} leans towards the diffusion prediction $\hat{\mathbf{x}}_0[i]$, while reliable areas are guided by the features from $\mathbf{g}[i]$. For simplicity, we let $\mathbf{p}$ denote corresponding positions in both image and latent space. In practice, $U_i$ is downsampled via average pooling to match the latent resolution before computing $\gamma_{t,i}$.
After computing $\gamma_{t,i}$, we apply Eq.~\eqref{eq:fusion_solution} to obtain the fused latent $\widetilde{\mathbf{x}}_0[i]$, which is then used in Eq.~\eqref{eq:denoise_detail_main} to update $\mathbf{x}_t$ to $\mathbf{x}_{t-1}$. This reverse sampling step is repeated until the final latent $\mathbf{x}_0$ is obtained, which is then decoded into pseudo-view images via the VAE decoder (see Fig.~\ref{fig:framework}).

\vspace{-0.5em}
\subsubsection{Extending to View Interpolation}\label{sec:extend_to_view_interp}
\vspace{-0.5em}
% The generation pipeline introduced above supports view extrapolation from one input view which may fail to keep the scene fidelity if the viewing angle deviates too far from the input view. 

The above generation pipeline supports view extrapolation from a single input, but may struggle to preserve scene fidelity under large viewpoint shifts. 
To alleviate this issue, we extend it to view interpolation using two input views as references. 
We define camera trajectories between them and run the diffusion model forward and backward, conditioned on the start and end images, respectively. At each denoising step, we merge the two latent sequences
 % Specifically, we define pseudo-view camera trajectories between two input views and run the diffusion model both forward and backward, conditioned on the start and end images, respectively. Then we merge the resulting latents of each denoising step during reverse sampling:
% To make the generated pseudo-views start from one input and end at another 
% To better align the generation with the real scene, we define camera trajectories for view generation starting from one input image and ending at another. 
% Concretely, after running the diffusion model both forward and backward, using the start and end images respectively, we merge the two resulting latents during the reverse sampling process: 
$\mathbf{x}_{t-1} := \bm{\beta} \mathbf{x}_{t-1}^{\text{forward}}+ (1-\bm{\beta}) R(\mathbf{x}_{t-1}^{\text{backward}})$, where $R(\cdot)$ is the reverse operation along the frame index dimension to align the latent $\mathbf{x}_{t-1}^{\text{backward}}$ to $\mathbf{x}_{t-1}^{\text{forward}}$ in the frame dimension. $\bm{\beta} \in \mathbb{R}^N$ is the blending weight, with $\bm{\beta}[i] = (N - i)/(N - 1)$ for $i = 1, 2, \ldots, N$, where $N$ is number of interpolated frames between two inputs. See supplementary material for the detailed algorithm. 

\vspace{-0.5em}
\subsection{3D-GS Optimization Guided by Generation}\label{sec:refinementa}
\vspace{-0.5em}
% Given sparse inputs of $M$ images along with their poses, 
% we aim at optimizing a 3DGS model with the auxiliary generated pseudo views by the video diffusion model.
% between the inputs. 
% For simplicity, we refer to the input images paired with their poses as ‘input views’, and we term the generated images with their associated poses as ‘generated views’.
% To better constrain the 3D-GS, 
To constrain the 3D-GS representation, we generate pseudo views by pairing adjacent input images and defining camera trajectories that better cover under-observed regions (see supplementary materials for details).
Using the video diffusion model guided by the generated guidance images  $\mathcal{I}^{\guid}$, 
% =\{{I^{\guid}_i}\}_{i=1}^{N}$,  
as described in Sec.~\ref{sec:pseudo_view_generation}, we interpolate between input views to obtain pseudo-view images 
$\mathcal{I}^{\pse} = \{I^{\pse}_j\}_{j=1}^{pN}$. 
where $p$ is the number of input image pairs.

\vspace{-0.5em}
\subsubsection{Gaussian Primitive Densification} \label{sec:gs_densification}
\vspace{-0.3em}
% Sparse-input 3D-GS training often leads to poor Gaussian-primitive reconstruction in under-observed regions due to limited supervision. 
Sparse-input 3D-GS training often yields poor reconstruction in under-observed regions due to limited supervision. 
To mitigate this, we enhance 3D-GS geometry using generated pseudo-views $\mathcal{I}^{\pse}$ and a dense stereo model~\cite{wang2024dust3r}.
For efficiency, we select a subset of pseudo-views $\mathcal{I}^{\den}\subseteq \mathcal{I}^{\pse}$ whose camera poses yield low inter-frame covisibility, ensuring broad scene coverage with minimal redundancy. 
These views are used to build a camera graph and optimize a point cloud from stereo predictions.
To further improve robustness, we analyze the spatial distribution of the reconstructed points and filter out those that significantly deviate from the global average distance to neighboring points.
Finally, we query existing Gaussian primitives within a fixed radius of each remaining point and only add new Gaussian primitives at positions without nearby primitives to augment the current set. See appendix for more details. 
% {\color{blue}See supplementary for more details.} 
% to avoid redundancy.
% After obtaining the optimized scene point cloud, we query the existing Gaussian primitives within a specified radius around each point. To avoid redundancy, only points that lack nearby Gaussian primitives are used to complement and expand the current primitive set.
% across the entire point cloud.
% To mitigate outliers caused by erroneous depth estimates, we analyze the spatial distribution of neighboring points and remove those that significantly deviate from others.
% the global average across the entire point cloud.

% After having the optimized scene point clouds, we query the Gaussian primitives that is within a specified radius around each point and only use the ones without neighboring Gaussian primitatives to complement the current primitive set. 

% \subsection{3D Gaussian Splatting Refinement} \label{sec:refinementa}
% \vspace{-0.5em}
\subsubsection{3D Gaussian Splatting Optimization} \label{sec:refinement}
% \vspace{-0.5em}

% We optimize the 3D-GS by including these generated pseudo views for training.  In each iteration, we sample an input view and a generated pseudo-view for supervision. 
% After densifying the Gaussian primitive set, 
% we optimize the 3D-GS model, also incorporating the generated pseudo-views into training. 
After densifying the Gaussian primitive set, we optimize the 3D-GS model using both the original inputs and the generated pseudo-views.
% In each 3D-GS training iteration, we sample one input view and one pseudo-view to provide supervision.
In each training iteration, one input view and one pseudo-view are sampled for supervision.
For the original input views, we apply an L1 loss and a D-SSIM loss
, as well as a depth regularization term $\mathcal{L}_\text{reg}$ with Pearson correlation similar to \cite{wang2025use}: $\mathcal{L}_s = w_1 \mathcal{L}_1(I^{\gs}, I^{\inp})+ w_2 \mathcal{L}_\text{D-SSIM}(I^{\gs}, I^{\inp}) + w_3 \mathcal{L}_\text{reg}$, where $I^{\gs}$ is the rendered image from 3D-GS and $I^{\inp}$ denotes the corresponding input image.
For the generated pseudo views, we observe that, despite the carefully designed guidance mechanism, some regions still suffer from temporal inconsistency---particularly distant areas with weak geometry or those with fine-grained textures, \eg, grass or tree leaves. 
To mitigate the negative impact of such inconsistencies on 3D-GS training, we use the LPIPS loss~\cite{zhang2018perceptual} instead of L1 loss. 
% The loss function for pseudo-views is given by:
The resulting loss for pseudo-views is:
% The loss function for the interpolated views is thus expressed as: 
\begin{equation}\label{eq:loss}
   \mathcal{L}_g = w_4 \mathcal{L}_\text{LPIPS}(I^{\gs},I^{\pse})+ w_5 \mathcal{L}_\text{D-SSIM}(I^{\gs}, I^{\pse})+ w_6 \mathcal{L}_\text{reg}. 
\end{equation}

\begin{figure*}[t]
    \centering
    \resizebox{0.99\linewidth}{!}{
    \includegraphics[page=8, angle=0, width=1\linewidth, trim={0.5cm 8cm 7cm 1cm},clip]{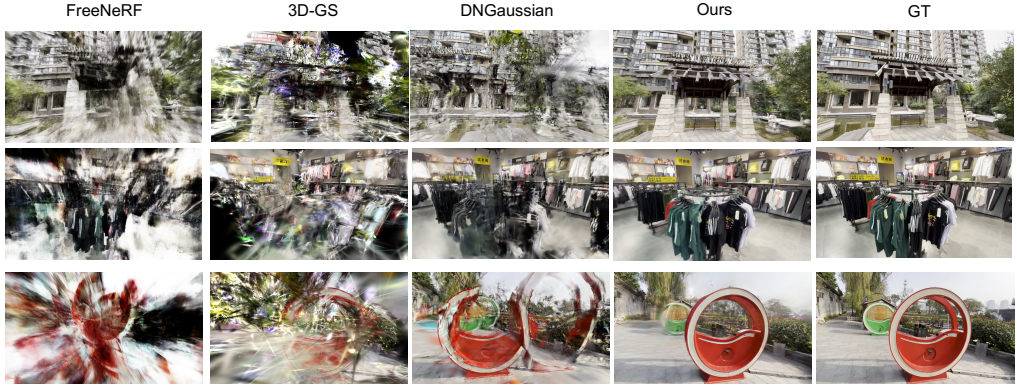}
    }
    % \vspace{-1em}
    \caption{
    % Qualitative comparison with other counterpart methods on DL3DV dataset. Thanks to the constraints from video diffusion, our method synthesizes photorealistic novel views under sparse inputs (9 views) while other counterparts generate noisy renderings. 
    \textbf{Qualitative comparison with existing methods on the DL3DV dataset} demonstrates the robustness of our methods against sparse inputs. 
    Leveraging the priors of the video diffusion model, our method renders photorealistic novel views from only 9 input views, while other methods produce noisier, less realistic results.
    % Our approach produces photorealistic novel views under sparse input conditions (9 views), benefiting from the strong priors and constraints imposed by the video diffusion model. In contrast, other methods struggle with the limited supervision and yield noticeably noisier or less realistic renderings.
    }
    \label{fig:qualitative_dl3dv}
    \vspace{-0.5em}
\end{figure*}

% \subsubsection{Guidance image creation}

% \subsubsection{Uncertainty evaluation }%\label{sec:uncertainty}

% \subsubsection{Uncertainty-aware Modulation}%\label{sec:reverse_sample_modulation}

% \subsection{3D-GS Optimization Guided by Generation}%\label{sec:refinementa}

% \subsubsection{Scene point cloud densification}
% % \textbf{Pseudo view filtering}

% \subsubsection{loss}

% \begin{figure}[t]
%     \centering
%     \includegraphics[page=3, width=0.9\linewidth, trim={0cm 6cm 9cm 0cm},clip]{fig/diagrams.pdf}
%     \vspace{-1em}
%     \caption{
%     % Illustration of the 3D-GS enhancement for pseudo view generation.
%     Uncertainty estimation for pseudo-view interpolation. The incorrect regions in the 3D-GS rendering (during training) are reliably identified by cross-view consistency check in terms of the photometric and geometric consistency.  
%     % The uncertainty estimation  
%     }
%     \label{fig:geo_photo_unceratinty}
%     % \vspace{-4em}
% \end{figure}

\vspace{-1em}
\section{Experiments}\label{sec:experiment}
\vspace{-0.5em}

\subsection{Experiment Settings}\label{sec:exp settings}
\vspace{-0.3em}

\textbf{Datasets and Metrics}.
We evaluate our method on LLFF \cite{llff}, DL3DV~\cite{ling2024dl3dv}, DTU \cite{dtu}, and MipNeRF-360~\cite{barron2022mip} datasets. 
LLFF consists of 8 forward-facing scenes. Following standard practice~\cite{yang2023freenerf, li2024dngaussian}, we train our model using only 3 input views on this dataset. 
DL3DV comprises diverse indoor and outdoor scenes, captured by humans walking through scenes. The Mip-NeRF 360 dataset consists of real-world indoor and outdoor scenes designed for evaluating novel view synthesis in large, unbounded environments. 
% Unlike bounded object-centric datasets, it features complex lighting, depth variations, and distant backgrounds, posing challenges for models to handle multi-scale geometry and view-dependent effects.
Compared to LLFF, DTU, and Mip-NeRF 360, DL3DV offers more diverse scene types and exhibits significantly more complex and dynamic camera motions. We include this dataset to evaluate the robustness of our approach under more realistic and challenging conditions, and our evaluation is conducted on the official test split of DL3DV. 
To verify the generalizability of our methods and compare with the previous methods, we also test our methods on DTU, an object-centric dataset captured in controlled conditions. For the DTU dataset, we follow the protocol from RegNeRF~\cite{li2024dngaussian}, using 3 training views (IDs 25, 22, and 28) across 15 evaluation scenes. To focus on the object of interest, we mask out the background during evaluation using the provided object masks, consistent with~\cite{yang2023freenerf, li2024dngaussian}.
We apply a downsampling factor of 8 for LLFF and 4 for DTU, aligning with prior work. The rendering quality is assessed using PSNR, SSIM, and LPIPS metrics.
% Aligning with the protocol of baselines, we apply the downsampling rate of 8 and 4 on the LLFF and DTU datasets, respectively. We evaluate the rendering quality using SSIM, LPIPS, and PSNR. 

\textbf{Implementation details}. 
% We implement our method with 3D Gaussian Splatting b. 
Our pipeline is designed to operate iteratively. In each cycle, we train the 3D-GS model for 10K iterations, followed by an update of the pseudo-view images using the video diffusion model. After each pseudo-view update, we reset the learning rate schedule of 3D-GS before starting the next optimization cycle to avoid overfitting. 
% 3D-GS optimizatio in the subsequent cycle.
For the uncertainty estimation in Eq.~\eqref{eq:uncertainty}, we set the bandwidth parameters to $s_1 = 100$ and $s_2 = 0.25$. The $\delta$ in Eq.~\eqref{eq:gamma} is fixed at $0.5$ across all experiments. The loss weights are configured as follows: $w_1 = 0.8$, $w_2 = 0.2$, $w_3 = 1.0$, $w_4 = 1.0$, $w_5 = 0.2$, and $w_6 = 1.0$.
% We use DPT \cite{Ranftl2021} to predict monocular depth maps for regularization in $\mathcal{L}_{reg}$. 
Additional implementation details are provided in supplementary materials. % For depth regularization in $\mathcal{L}_{reg}$, we utilize monocular depth maps predicted by DPT~\cite{Ranftl2021}. All experimental results are obtained on a single A40 GPU. Additional implementation details are provided in the supplementary materials.

\begin{figure*}[t]
    \centering
    \resizebox{1\linewidth}{!}{
    \includegraphics[page=19, angle=0, width=1\linewidth, trim={0cm 9cm 2.7cm 0cm},clip]{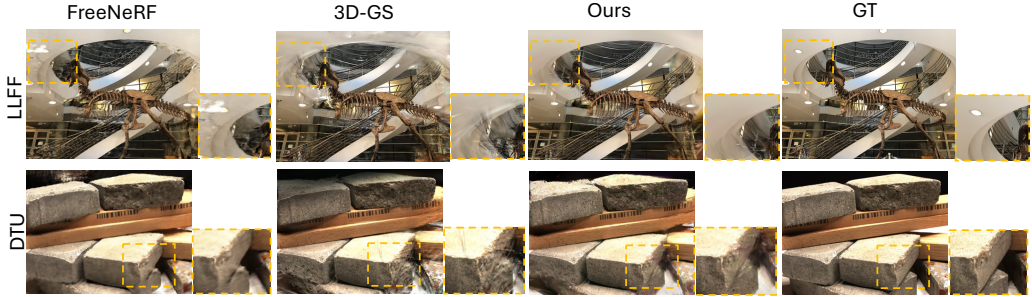}
    }
    \vspace{-2em}
    \caption{
    Qualitative comparison with other methods on DTU and LLFF datasets. 
    }
    \label{fig:qualitative_llff_dtu}
    \vspace{-1em}
\end{figure*}

\begin{table}[t]
\centering
\caption{
\textbf{Quantitative comparisons with other methods on LLFF, DTU, and MipNeRF-360} demonstrate our state-of-the-art performance and strong generalization ability.
\textbf{Left}:  Quantitative comparison with other methods on the LLFF dataset with 3 training views. 
\textbf{Middle}: Comparison on the DTU dataset with 3 training views. 
\textbf{Right}: Comparison on the MipNeRF-360 dataset with 9 training views. Recent reconstruction-based methods, feed-forward methods, and non-zero-shot methods are included.
% \textbf{Right}: Quantitative comparison with other methods on the DL3DV dataset with 3, 6 and 9 training views.
We color each cell as \textbf{\colorbox[RGB]{255,179,179}{best}}, \textbf{\colorbox[RGB]{255,217,179}{second best}}, and \textbf{\colorbox[RGB]{255,255,179}{third best}}. \label{tab:main_table}
}

\begin{subtable}[t]{0.31\textwidth}
\centering
\resizebox{1\linewidth}{!}{
\begin{tabular}{r|ccc} \toprule
\textbf{\textit{LLFF}}             & PSNR$\uparrow$ & SSIM$\uparrow$ & LPIPS$\downarrow$      \\ \hline
Mip-NeRF~\cite{barronmip}            & 16.11              & 0.401              & 0.460                            \\
3D-GS~\cite{kerbl20233d}              & 17.43              & 0.522              & 0.321                        \\ \hline
DietNeRF~\cite{jain2021putting}           & 14.94             & 0.370               & 0.496                          \\
RegNeRF~\cite{niemeyer2022regnerf}            & 19.08              & 0.587              & 0.336                           \\
FreeNeRF~\cite{yang2023freenerf}           & 19.63             & 0.612               & 0.308                            \\
SparseNeRF~\cite{wang2023sparsenerf}         & 19.86             & 0.624               & 0.328                           \\

\hline
{FSGS}~\cite{zhu2025fsgs}        & \cellcolor{top3}{20.31}   & \cellcolor{top3}{0.652}  & \cellcolor{top3}{0.288}     \\ 
{DNGaussian}~\cite{li2024dngaussian}   & {19.12}                   & {0.591}   & {0.294}     \\ 
IPSM~\cite{wang2025use} &\cellcolor{top2}20.44 & \cellcolor{top2}0.702 &  \cellcolor{top2}0.207     \\
\textbf{Ours}   & \cellcolor{top1}{20.61}  & \cellcolor{top1}{0.705} & \cellcolor{top1}{0.201}    \\ 
\bottomrule
\end{tabular}\label{tab:llff}
}
\end{subtable}
\hfill\hfill
\begin{subtable}[t]{0.3\textwidth}
 \resizebox{1\linewidth}{!}{
    \begin{tabular}{r|ccc} \toprule
    \multirow{1}{*}{ \textbf{\textit{DTU}}} & PSNR$\uparrow$ & SSIM$\uparrow$ & LPIPS$\downarrow$   \\ \hline
    Mip-NeRF~\cite{barronmip}  &8.68   &0.571   &  0.353   \\
    3D-GS~\cite{kerbl20233d}     &10.99  &0.585 &0.313              \\ \hline
    DietNeRF~\cite{jain2021putting}  &11.85  &0.633 &0.314                    \\
    RegNeRF~\cite{niemeyer2022regnerf}   &18.89  &0.745 &0.190          \\
    FreeNeRF~\cite{yang2023freenerf}  &\cellcolor{top2} 19.92  &0.787 &0.182            \\
    SparseNeRF~\cite{wang2023sparsenerf} &\cellcolor{top3}19.55  &0.769 & 0.201     \\
    \hline
    % {FSGS}                              &           &          &          &           &         \\ 
    {DNGaussian}~\cite{li2024dngaussian} &18.91  &\cellcolor{top3}0.790  &\cellcolor{top2}0.176    \\ 
    % {SparseGS} &   18.89 & 0.702 & 0.229    \\ 
    {SparseGS}~\cite{xiong2023sparsegs} &   18.89 & \cellcolor{top2}0.834 & \cellcolor{top3}0.178    \\ 
    \textbf{Ours} & \cellcolor{top1}20.51 & \cellcolor{top1}0.840  &\cellcolor{top1} 0.137  \\ 
    \bottomrule
    \end{tabular}
    \label{tab:dtu}
}
\end{subtable}
\hfill
\begin{subtable}[t]{0.35\textwidth}
    
 \resizebox{1\linewidth}{!}{
    \begin{tabular}{r|ccc} \toprule
    \multirow{1}{*}{ \textbf{\textit{MipNeRF-360}}} & PSNR$\uparrow$ & SSIM$\uparrow$ & LPIPS$\downarrow$   \\ \hline
    % Mip-NeRF  &8.68   &0.571   &  0.353   \\
    % 3D-GS     &10.99  &0.585 &0.313              \\ \hline
    % DietNeRF  &11.85  &0.633 &0.314                    \\
    RegNeRF~\cite{niemeyer2022regnerf}   &13.73  &0.193 &0.629          \\
    FreeNeRF~\cite{yang2023freenerf}  &13.20  &0.198 &0.635            \\
    DNGaussian~\cite{li2024dngaussian} &12.51                   &0.228                  &0.683\\ \hline
    MVSplat 360~\cite{chen2024mvsplat360} &14.86                  &0.321                  &\cellcolor{top3} 0.528 \\
    ViewCrafter~\cite{yu2024viewcrafter}	& \cellcolor{top2} 16.68 & \cellcolor{top3}0.382 & 0.551 \\
    3DGS-Enhancer~\cite{liu3dgs} &\cellcolor{top3}16.22 & \cellcolor{top2}0.399 & \cellcolor{top2}0.454 \\
    \hline
    \textbf{Ours} & \cellcolor{top1}17.91 & \cellcolor{top1}0.495  &\cellcolor{top1}0.435  \\ 
    \bottomrule
    \end{tabular}
    % \label{tab:mip360}
    }
\end{subtable}

\vspace{-1ex}

\end{table}

\begin{table}[t]
\centering
\caption{ \textbf{Our method outperforms other test-time optimization methods on the DL3DV dataset}. 
The results are reported for 3, 6, and 9 training views. We color each cell as \textbf{\colorbox[RGB]{255,179,179}{best}}, \textbf{\colorbox[RGB]{255,217,179}{second best}}, and \textbf{\colorbox[RGB]{255,255,179}{third best}}.  }\label{tab:dl3dv}

\resizebox{0.8\linewidth}{!}{%
\begin{tabular}{l|ccc|ccc|ccc} \toprule  
 & \multicolumn{3}{c}{\textbf{3 Views}} & \multicolumn{3}{|c|}{\textbf{6 Views}} & \multicolumn{3}{c}{\textbf{9 Views}} \\ \cline{2-10} 
 % \vspace{-1ex} \\
 {Method} & PSNR$\uparrow$ & SSIM$\uparrow$ & LPIPS$\downarrow$ & PSNR$\uparrow$ & SSIM$\uparrow$ & LPIPS$\downarrow$ & PSNR$\uparrow$ & SSIM$\uparrow$ & LPIPS$\downarrow$ \\ \midrule
 % \multicolumn{10}{c}{\textbf{DL3DV} (130 training scenes, 20 test scenes)} \\ \midrule 
 Mip-NeRF \cite{barronmip}     & 10.92 & 0.191 &  0.618  & 11.56 & 0.199 & 0.608  & 12.42 & 0.218 & 0.600  \\ 
 3DGS \cite{kerbl20233d}         & 10.97 &  0.248 &  0.567  &  12.34 &  0.332 &  0.598 & 12.99 &  0.403 &  0.546    \\\hline
 % DietNeRF \cite{niemeyer2022regnerf}    &  &  &  &  &  &  &  &   &   \\
 RegNeRF \cite{niemeyer2022regnerf}    & 11.46 & 0.214 & 0.600 & 12.69 & 0.236 & 0.579 & 12.33 &  0.219 &  0.598 \\
 FreeNeRF~ \cite{yang2023freenerf}         &  10.91 &  0.211 &  0.595  &  12.13 &  0.230 &  0.576 &  12.85 &  0.241 &  0.573    \\ \hline
 FSGS \cite{zhu2025fsgs} &   \cellcolor{top2} 12.22 & \cellcolor{top2}  0.296 & \cellcolor{top3} 0.535 &  \cellcolor{top2} 13.73 & \cellcolor{top2}0.429  & \cellcolor{top3} 0.540 & \cellcolor{top2} 15.52  & \cellcolor{top2}0.468  &  \cellcolor{top2} 0.416      \\
 DNGaussian \cite{li2024dngaussian} & 11.10 & 0.273 & 0.579 & 12.67 &  0.329 & 0.547 &  \cellcolor{top3} 13.44 & \cellcolor{top3}0.365 &  0.539    \\
 IPSM \cite{wang2025use} & \cellcolor{top3} 11.70 & \cellcolor{top3} 0.279 &  \cellcolor{top2} 0.534 & \cellcolor{top3} 12.82 &  \cellcolor{top3}0.332 & \cellcolor{top2} 0.521 &  13.41 & 0.361 &  \cellcolor{top3} 0.529    \\
 % 3DGSEnhancer$^\dagger$~\cite{liu3dgs} & \cellcolor{top2}14.33 &  \cellcolor{top3} 0.424 &  \cellcolor{top1}\cellcolor{top1} {0.464}  & \cellcolor{top2}{16.94}  &\cellcolor{top2}{ 0.565} &\cellcolor{top1}{0.356 } & \cellcolor{top2}{18.50}  & \cellcolor{top2}0.630 & \cellcolor{top1}{0.305} \\
 Ours  & \cellcolor{top1} 14.62  &  \cellcolor{top1}  0.471 & \cellcolor{top1}  0.491 &  \cellcolor{top1}17.35 &  \cellcolor{top1}0.566 &  \cellcolor{top1}0.396& \cellcolor{top1} 19.19 & \cellcolor{top1} 0.616 & \cellcolor{top1}0.335\\
 \bottomrule
\end{tabular}
} 
\vspace{-1ex}
\end{table}

\vspace{-0.5em}
\subsection{Comparison with Other Methods}
\vspace{-0.5em}
We compare our method against state-of-the-art approaches on four benchmark datasets to demonstrate its effectiveness and generalizability across diverse scenarios.

\noindent\textbf{Comparison on LLFF}. 
% To verify the generalizability, we further 
We evaluate our method on the LLFF dataset captured by a swaying face-forward camera. 
% The camera motion is relatively small compared with DL3DV. 
% As shown in Table~\ref{tab:main_table} (left), our method consistently outperforms the NeRF-based methods in all metrics. Compared with the 3D-GS-based counterparts FSGS~\cite{zhu2025fsgs} and DNGaussian~\cite{li2024dngaussian}, our performance is still competitive, particularly in LPIPS and SSIM metrics, thanks to the additional constraints from the frames interpolated by the video diffusion model. 
As shown in Table~\ref{tab:main_table} (left), our method consistently outperforms NeRF-based approaches across all evaluation metrics. When compared to 3D-wGaussian Splatting–based baselines such as FSGS~\cite{zhu2025fsgs} and DNGaussian~\cite{li2024dngaussian}, our method remains competitive, particularly in LPIPS and SSIM scores. This improvement is largely attributed to the additional supervisory signal provided by the pseudo views generated through the video diffusion model. Notably, the LPIPS metric, which correlates more closely with human perceptual similarity than traditional metrics like PSNR, highlights our method’s ability to produce visually realistic novel views. Qualitative comparisons are presented in Fig.~\ref{fig:qualitative_llff_dtu}.

% LPIPS metric aligns more closely with human perception compared to traditional metrics like PSNR. It shows that our method can synthesize realistic novel views that match humans' assessment. The qualitative results are shown in Fig.~\ref{fig:qualitative_llff_dtu}. 
\noindent\textbf{Comparison on DTU}. 
To further assess the generalizability of our approach, we evaluate and compare its performance on the DTU dataset. DTU is an object-centric dataset in which each scene contains a centered object against a monotone background.  The evaluation results are presented in Table~\ref{tab:main_table} (middle). 
In this setting, our method still performs well and outperforms other NeRF-based and 3D-GS-based methods.  Specifically, our method outperforms the second-best approach by a significant margin in terms of PSNR, SSIM, and LPIPS. 
% \ie, $20.51$ vs. $19.92$ on PSNR,  $0.840$ vs. $0.790$ on SSIM, $0.137$ vs. $0.176$ on LPIPS. 
While NeRF-based methods also exhibit competitive accuracy in this scenario, they suffer from slow rendering speeds (approximately 0.21 FPS), whereas our 3D-GS-based approach supports real-time rendering at around $430$ FPS.

\noindent\textbf{Comparison on DL3DV}.  
% DL3DV contains images captured by humans walking through the outdoor/indoor scenes. We choose this The camera motion in this dataset is more significant compared with the LLFF and DTU. 
% Unlike the synthetic datasets that most previous methods work on, this dataset is more complex, and the camera motion is more significant.   
We compare with other cutting-edge counterparts on the DL3DV dataset under 3,6, and 9 view settings. Table~\ref{tab:dl3dv} shows the quantitative comparison results. 
Apart from the sparse-input 3D-GS methods, we also compare with the non-sparse view methods and NeRF-based methods in  Table~\ref{tab:dl3dv}. 
We outperform previous state-of-the-art methods~\cite{li2024dngaussian, zhu2025fsgs, wang2025use} by a significant margin in this challenging setting. 
We observe that although DNGaussian~\cite{li2024dngaussian} works well in environments with limited scope or with limited camera motions, \eg, object-centric scenarios, it has difficulties in reliably reconstructing the open environment due to the lack of constraints in under-observed regions, as the sparse (a qualitative results shown in Fig.~\ref{fig:qualitative_dl3dv}). 
Similarly, FSGS~\cite{zhu2025fsgs} also struggles in this challenging setting, though it achieves slightly better performance compared with DNGaussian because it uses a sparse point cloud for initialization. The recent work IPSM~\cite{wang2025use} uses an image diffusion model to constrain the 3D-GS by enhancing Score Distillation Sampling (SDS). 
As shown in Table~\ref{tab:dl3dv}, this method struggles with extremely sparse inputs. This limitation arises because the image diffusion model lacks access to a global scene context, whereas the video diffusion model is able to infer such context from the input reference frame.

\begin{table}[t]
  \centering

% \begin{minipage}{0.28\linewidth}
%     \centering
%     \caption{
%     % Quantitative 
%     Comparison with other counterparts on the DTU with 3 training views. 
%     % Our method generalizes well to this object-level dataset, and outperforms other 3D-GS based methods while maintaining an efficient rendering speed. 
%     }
%     % \vspace{-0.5em}
%     \resizebox{1\linewidth}{!}{
%     \begin{tabular}{c|ccc} \toprule
%     \multirow{1}{*}{Methods} & PSNR$\uparrow$ & SSIM$\uparrow$ & LPIPS$\downarrow$   \\ \hline
%     Mip-NeRF  &8.68   &0.571   &  0.353   \\
%     3D-GS     &10.99  &0.585 &0.313              \\ \hline
%     DietNeRF  &11.85  &0.633 &0.314                    \\
%     RegNeRF   &18.89  &0.745 &0.190          \\
%     FreeNeRF  &\cellcolor{top2} 19.92  &\cellcolor{top3}0.787 &0.182            \\
%     SparseNeRF &\cellcolor{top3}19.55  &0.769 & \cellcolor{top3} 0.201     \\
%     \hline
%     % {FSGS}                              &           &          &          &           &         \\ 
%     {DNGaussian} &18.91  &\cellcolor{top2}0.790  &\cellcolor{top2}0.176    \\ 
%     % {FewViewGS} &    & &  &  &  \\ 
%     {SparseGS} &   18.89 & 0.702 & 0.229    \\ 
%     \textbf{Ours} & \cellcolor{top1}20.51 & \cellcolor{top1}0.840  &\cellcolor{top1} 0.137  \\ 
%     \bottomrule
%     \end{tabular}
%     \label{tab:dtu}
%     }

% \end{minipage}
% \hfill
\begin{minipage}{\linewidth}
\centering
\vspace{-1em}
\captionof{table}{\textbf{Ablation experiments on the DL3DV test set}.  
(a) Experiments to show the effectiveness of the proposed components in pseudo-view generation step. (b) Experiments to show the effectiveness of the proposed strategies for 3D-GS optimization.  } \label{tab:ablation}
\begin{subtable}[t]{0.51\textwidth}
\setlength{\tabcolsep}{0.8mm}{
\resizebox{\linewidth}{!}{
\begin{tabular}{l|ccc}
\toprule
% baseline w/o interp 
% guidance image creation -- 1) gs render 2) warpping
% uncertainty -- 1) w/ 2) w/o 3) w/o photometric 4) w/o geometric 
% \tau values
% 

(a) & \multirow{1}{*}{PSNR\textuparrow} & \multirow{1}{*}{SSIM\textuparrow}& \multirow{1}{*}{LPIPS\textdownarrow}   \\   
\hline \\ \\\vspace{-2.6em} \\
Baseline 3D-GS & 16.59 & 0.502 & 0.405 \\
w/ GS interpolation & 18.59 & 0.591 &0.369\\
% w/ warping interpolation (full)  &\textbf{19.16} &\textbf{0.631} &\textbf{0.341} \\
w/ warping interpolation (full)  &\textbf{19.19} &\textbf{0.616} &\textbf{0.335} \\
~~~~w/o geometric &18.21 & 0.583 & 0.378\\
~~~~w/o photometric &18.93 & 0.612 & 0.352 \\

\bottomrule
\end{tabular}
}
}
\end{subtable}
\hfill
\begin{subtable}[t]{0.46\textwidth}
\setlength{\tabcolsep}{0.8mm}{
\resizebox{\linewidth}{!}{
\begin{tabular}{l|ccc}
\toprule
%$\mathcal{L}_{reg}$  & L1 & LPIPS& Uncertainty & %$\mathcal{L}_{reg}$  & L1 & LPIPS
(b)& \multirow{1}{*}{PSNR\textuparrow} & \multirow{1}{*}{SSIM\textuparrow}& \multirow{1}{*}{LPIPS\textdownarrow}   \\  
\hline \\\vspace{-1.6em} \\  
w/o point filtering  &19.01 &0.615 &0.343 \\
w/o GS densification &18.23 &0.567 &0.386 \\
w/o LPIPS loss &18.81 &0.597& 0.351 \\
% w/o $\mathcal{L}_{reg}$ &18. & &\\
Full model & \textbf{19.19} & \textbf{0.616} & \textbf{0.335}\\
\bottomrule
\end{tabular}
}
}
\end{subtable}

% \end{minipage}
% \begin{minipage}{0.5\textwidth}
%     \centering
%     % \resizebox{0.99\linewidth}{!}{
%     \includegraphics[page=25, angle=0, width=0.97\linewidth, trim={0.5cm 5cm 16cm 0.5cm},clip]{fig/diagrams.pdf}
%     \captionof{figure}{Qualitative comparison with the feed-forward method trained with large-scale data. Our method better keeps the visual/geometric consistency. }
%     % }
%     % \vspace{-0.5em}
%     \label{fig:mipnerf-360}
\end{minipage}
\vspace{-4ex}
\end{table}

\begin{figure}
\vspace{-5ex}
    \centering
    \includegraphics[page=25, angle=0, width=0.97\linewidth, trim={0.5cm 10cm 0.5cm 0cm},clip]{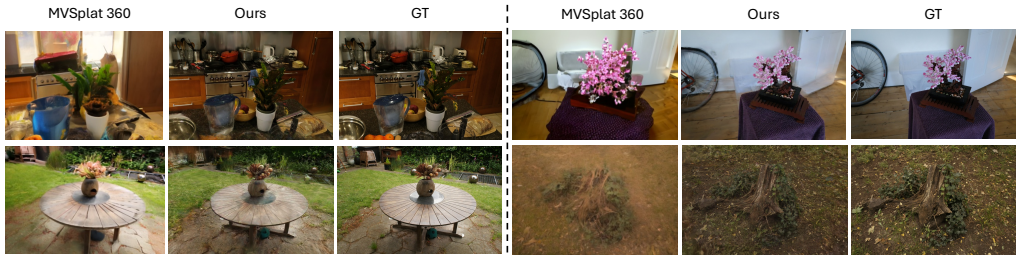}
    \caption{
\textbf{Our test-time optimization better preserves visual and geometric consistency than the feed-forward approach, MVSplat360}. While feed-forward methods can produce plausible novel views, they often struggle to maintain fidelity to the original scene, whereas our method achieves higher consistency.
    % and   
    % Qualitative comparison with the feed-forward method trained with large-scale data. Our method better keeps the visual/geometric consistency. 
    }
    \label{fig:mipnerf-360}
\end{figure}

\noindent\textbf{Comparison on MipNeRF-360}. To evaluate our method on unbounded scenes and ensure a fair comparison with recent feed-forward approaches~\cite{chen2024mvsplat360,yu2024viewcrafter,liu3dgs}, we further conduct experiments on the Mip-NeRF 360 dataset~\cite{barron2022mip}. Our method consistently outperforms reconstruction-based methods~\cite{niemeyer2022regnerf,yang2023freenerf,li2024dngaussian} and surpasses state-of-the-art feed-forward approaches~\cite{chen2024mvsplat360,yu2024viewcrafter,liu3dgs} by a notable margin. As shown in Fig.~\ref{fig:mipnerf-360}, although feed-forward methods can hallucinate novel views from sparse inputs through large-scale data training, they often struggle to maintain geometric consistency, fine details, and color fidelity compared to our approach.

% \begin{table}[]
%     \centering
%     \begin{minipage}{0.28\linewidth}
%     \centering
%     \caption{
%     % Quantitative 
%     Comparison with other counterparts on the MipNeRF-360 with 9 training views. 
%     % Our method generalizes well to this object-level dataset, and outperforms other 3D-GS based methods while maintaining an efficient rendering speed. 
%     }
%     % \vspace{-0.5em}
%     \resizebox{1\linewidth}{!}{
%     \begin{tabular}{c|ccc} \toprule
%     \multirow{1}{*}{Methods} & PSNR$\uparrow$ & SSIM$\uparrow$ & LPIPS$\downarrow$   \\ \hline
%     % Mip-NeRF  &8.68   &0.571   &  0.353   \\
%     % 3D-GS     &10.99  &0.585 &0.313              \\ \hline
%     % DietNeRF  &11.85  &0.633 &0.314                    \\
%     RegNeRF~\cite{niemeyer2022regnerf}   &13.73  &0.193 &0.629          \\
%     FreeNeRF~\cite{yang2023freenerf}  &13.20  &0.198 &0.635            \\
%     DNGaussian~\cite{li2024dngaussian} &12.51                   &0.228                  &0.683\\ \hline
%     MVSplat 360~\cite{chen2024mvsplat360} &14.86                  &0.321                  &\cellcolor{top3} 0.528 \\
%     ViewCrafter~\cite{yu2024viewcrafter}	& \cellcolor{top3} 15.68 & \cellcolor{top3}0.382 & 0.551 \\
%     3DGS-Enhancer~\cite{liu3dgs} &\cellcolor{top2}16.22 & \cellcolor{top2}0.399 & \cellcolor{top2}0.454 \\
%     \hline
%     \textbf{Ours} & \cellcolor{top1}17.91 & \cellcolor{top1}0.495  &\cellcolor{top1}0.435  \\ 
%     \bottomrule
%     \end{tabular}
%     \label{tab:mip360}
%     }

% \end{minipage}

% \end{table}

\begin{figure*}[t]
    % \vspace{-0.5ex}
    \centering
    % \resizebox{0.99\linewidth}{!}{
    \includegraphics[page=17, angle=0, width=0.97\linewidth, trim={0.5cm 13cm 11cm 0cm},clip]{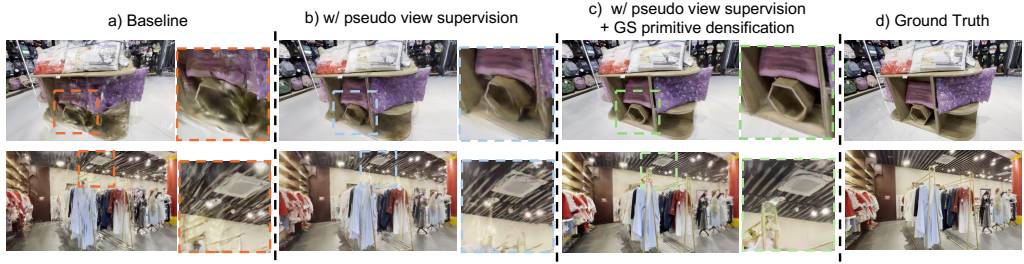}
    % }
    % \includegraphics[page=17, angle=0, width=0.45\linewidth, trim={0.5cm 8cm 6.5cm 1cm},clip]{fig/diagrams.pdf}
    \vspace{-0.5em}
    \caption{
    % Qualitative results to show the effectiveness of our proposed components. 
    \textbf{The proposed pseudo-view supervision and primitive densification effectively enhance the novel view synthesis}, especially in under-observed regions from the inputs.  
    % The pseudo-view supervision and the GS primitive densification enhance the 
    % training in under-observed regions, leading to more realistic novel view synthesis.  
    Zoom in for a better view. 
    }
    \label{fig:qualitative_ablation}
    \vspace{-1.em}
\end{figure*}

\begin{figure*}[t]
    \centering
    % \resizebox{0.99\linewidth}{!}{
    \includegraphics[page=18, angle=0, width=1\linewidth, trim={0.5cm 14cm 0cm 0cm},clip]{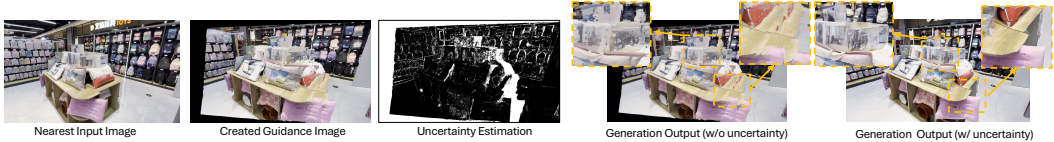}
    % }
    % \includegraphics[page=17, angle=0, width=0.45\linewidth, trim={0.5cm 8cm 6.5cm 1cm},clip]{fig/diagrams.pdf}
    \vspace{-2em}
    \caption{
    \textbf{The estimated uncertainty mask identifies the unreliable parts in guidance images. }
    The video diffusion model cannot generate faithful images without involving the uncertainty mask.   
    }
    \label{fig:uncertain_svd}
    \vspace{-1.5em}
\end{figure*}

\subsection{Ablation Study}
\vspace{-0.5em}
To validate the effectiveness of our proposed components in the pseudo view generation (Sec.~\ref{sec:pseudo_view_generation}) and the 3D-GS optimization (Sec.~\ref{sec:refinementa}), we conduct an extensive ablation study on DL3DV. 
% In Fig.~\ref{fig:qualitative_ablation}

% \noindent\textbf{Effectiveness of uncertainty-aware reverse sampling modulation.}

% We first evaluate the performance using the renderings from 3D-GS as the guidance view to guide the video diffusion model (`w/ GS interpolation'). 

\noindent\textbf{Effectiveness of uncertainty-aware modulation mechanism.}
Table~\ref{tab:ablation}a compares the baseline 3D-GS trained on sparse views using $\mathcal{L}_{s}$ with two variants: one using 3D-GS renderings as guidance (“w/ GS interpolation”) and one using our warping-based guidance (“w/ warping interpolation”). While GS interpolation improves over the baseline, it underperforms compared to our method due to inaccurate color rendering at novel poses during training.

%\noindent\textbf{Effectiveness of uncertainty-aware modulation mechanism.}i
%In Table~\ref{tab:ablation}(a), the baseline 3D-GS model is trained on sparse input views using the loss function $\mathcal{L}_{s}$.    Building on this baseline, we compare the performance of using 3D-GS renderings as guidance views (denoted as "w/ GS interpolation") and using our proposed guidance view creation approach (``w/ warping interpolation'') for diffusion guidance.  While the variant using 3D-GS renderings as guidance yields better performance than the baseline, it remains inferior to our proposed guidance image generation---inversely warping from the nearest input view. The performance gap is primarily due to the inaccurate color rendering of 3D-GS at novel poses during training.

% Despite better performance compared with the baseline, it is inferior compared to using our proposed guidance image creation procedure---inversely warping from the nearest input view (denoted as `w/ warping interpolation'). 

Fig.~\ref{fig:uncertain_svd} shows the effect of uncertainty-aware modulation by comparing diffusion results with and without it, using identical guidance images. We further ablate the geometric and photometric terms in the uncertainty formulation (Eq.~\eqref{eq:uncertainty}), denoted as “w/o geometric” and “w/o photometric.” As shown in Table~\ref{tab:ablation}a, removing either term noticeably degrades performance.
% To better illustrate the impact of our uncertainty estimation on the diffusion output, in Fig.~\ref{fig:uncertain_svd}, we compare the diffusion model outputs \red{w/ and w/o the uncertainty-aware modulation}. 

\noindent\textbf{Effectiveness of Gaussian primitive densification.}
We ablate the densification step (“w/o GS densification” in Table~\ref{tab:ablation}b), observing a significant performance drop, highlighting its role in improving synthesis quality. Fig.~\ref{fig:qualitative_ablation} shows that densification enhances reconstruction in under-observed regions. Removing the point filtering step (“w/o point filtering”) also degrades performance due to depth outliers from the stereo model.

% To verify the effectiveness of the proposed Gaussian primitive densification, we evaluate a pipeline by ablating the densification step from the full model (denoted as `w/o GS densification' in Table~\ref{tab:ablation}(b)). A significant performance drop occurs.  In Fig.~\ref{fig:qualitative_ablation}, we qualitatively compare the novel view synthesis results w/ or w/o GS primitive densification. It is shown that the proposed densification module can consistently improve the reconstruction in under-observed regions and lead to more faithful synthesis results. 
% We also tried to remove the point filtering step (`w/ point filtering') and found a performance degradation caused by the depth outliers from the stereo model   
%We ablate the densification step from the full pipeline (denoted as "w/o GS densification" in Table~\ref{tab:ablation}(b)). This results in a significant performance drop, demonstrating the importance of densification in improving the overall synthesis quality. Fig.~\ref{fig:qualitative_ablation} presents a qualitative comparison of novel view synthesis results with and without GS primitive densification. As shown, our densification module consistently enhances reconstruction in under-observed regions, leading to more faithful and complete synthesis. Additionally, we assess the impact of removing the point filtering step ("w/o point filtering") and observe a degradation in performance due to the depth outliers introduced by the stereo model.

\noindent\textbf{Effectiveness of LPIPS for pseudo view supervision.}
We replace LPIPS with L1 loss (“w/o LPIPS loss”) in Eq.~\eqref{eq:loss}, observing a notable performance drop (Table \ref{tab:ablation}b). Despite our guidance strategy, cross-view inconsistencies -- especially in distant or textured regions -- remain challenging. L1 loss used in vanilla 3D-GS~\cite{kerbl20233d} is less robust to such inconsistencies in diffusion-generated pseudo views.

\vspace{-1em}
\section{Conclusion and Limitation}\label{sec: conclusion}
\vspace{-1em}
% {\bf Summary.}
We introduced a zero-shot, generation-guided pipeline that leverages a pretrained video diffusion model to improve 3D-GS reconstruction from sparse inputs. Intermediate views are synthesized and guided by warped depth-based images and uncertainty-aware modulation. A densification module further enhances scene completeness. Our approach improves photorealism and coverage in sparse settings while maintaining the real-time efficiency of 3D-GS.

Our framework improves sparse-view synthesis but has limitations. It depends on the quality of the pretrained video diffusion model, which may introduce artifacts under extreme views or complex scenes. Iterative training adds overhead compared with vanilla 3D-GS pipelines, and early 3D-GS depth errors can affect guidance quality despite uncertainty modeling, though this impact typically decreases over time.

\textbf{Societal Impact.} 
This technology can benefit AR/VR, robotics, digital content creation, telepresence, and cultural heritage preservation. However, its computational demands may contribute to a higher carbon footprint.

\textbf{Acknowledgment.} This project was supported, in part, by NSF 2215542, NSF 2313151, and Bosch gift funds to S. Yu at UC Berkeley and the University of Michigan. 

% This technology has the potential to benefit applications in augmented and virtual reality (AR/VR), robotics, digital content creation, telepresence, and cultural heritage preservation. As with many generative models, our approach may be misused for creating synthetic or manipulated 3D scenes, raising concerns around misinformation, deepfakes, or privacy in surveillance applications. Additionally, diffusion-based models can be computationally intensive, contributing to increased carbon footprints.
%While our framework substantially improves novel view synthesis under sparse inputs, it has several limitations. First, our method depends on the quality of the pre-trained video diffusion model, which may introduce artifacts under extreme viewpoint shifts or in complex scenes. Second, the iterative training process adds computational overhead compared to vanilla 3D-GS pipelines.  Additionally, early-stage depth errors from sparse 3D-GS may affect guidance quality despite uncertainty modeling, although this impact tends to diminish as training progresses.

% Lastly, the approach assumes known and accurate camera poses, limiting applicability in pose-uncertain scenarios.

\medskip

{
\small
\bibliographystyle{plain}
\bibliography{main}
}

\end{document}